\documentclass[10pt,twocolumn,letterpaper]{article}

\usepackage[pagenumbers]{cvpr} 

\usepackage{amsmath}
\usepackage{amssymb}
\usepackage{graphicx,verbatimbox}

\usepackage{xspace}
\usepackage{booktabs}
\usepackage{multirow}
\usepackage{rotating}
\usepackage{color}
\usepackage{tablefootnote}
\usepackage{caption}

\renewcommand{\eg}{\emph{e.g.}}
\renewcommand{\etal}{\emph{et al.}}
\renewcommand{\ie}{\emph{i.e.}}

\newcommand{\F}{\mathcal{F}}

\newcommand{\R}{\mathbb{R}}

\newcommand{\tabincell}[2]{\begin{tabular}{@{}#1@{}}#2\end{tabular}}

\usepackage[pagebackref,breaklinks,colorlinks]{hyperref}

\usepackage[capitalize]{cleveref}
\crefname{section}{Sec.}{Secs.}
\Crefname{section}{Section}{Sections}
\Crefname{table}{Table}{Tables}
\crefname{table}{Tab.}{Tabs.}

\begin{document}

\title{Hire-MLP: Vision MLP via Hierarchical Rearrangement}

\author{
    Jianyuan Guo\textsuperscript{\rm 1,3}\thanks{Equal contribution.},
    Yehui Tang\textsuperscript{\rm 1,2*},
	Kai Han\textsuperscript{\rm 1},
	Xinghao Chen\textsuperscript{\rm 1},\\
	Han Wu\textsuperscript{\rm 3},
	Chao Xu\textsuperscript{\rm 2},
	Chang Xu\textsuperscript{\rm 3}\thanks{Corresponding author.},
	Yunhe Wang\textsuperscript{\rm 1$^\dagger$}
	 \\
	\textsuperscript{\rm 1}Noah’s Ark Lab, Huawei Technologies.\\
	\textsuperscript{\rm 2}Key Lab of Machine Perception (MOE), Dept. of Machine Intelligence, Peking University.\\
	\textsuperscript{\rm 3}School of Computer Science, Faculty of Engineering, University of Sydney.\\
	\{jianyuan.guo, kai.han, yunhe.wang\}@huawei.com, c.xu@sydney.edu.au.
}

\maketitle

\begin{abstract}
Previous vision MLPs such as MLP-Mixer and ResMLP accept linearly flattened image patches as input, making them inflexible for different input sizes and hard to capture spatial information. Such approach withholds MLPs from getting comparable performance with their transformer-based counterparts and prevents them from becoming a general backbone for computer vision.
This paper presents Hire-MLP, a simple yet competitive vision MLP architecture via \textbf{Hi}erarchical \textbf{re}arrangement, which contains two levels of rearrangements. Specifically, the inner-region rearrangement is proposed to capture local information inside a spatial region, and the cross-region rearrangement is proposed to enable information communication between different regions and capture global context by circularly shifting all tokens along spatial directions.
Extensive experiments demonstrate the effectiveness of Hire-MLP as a versatile backbone for various vision tasks. In particular, Hire-MLP achieves competitive results on image classification, object detection and semantic segmentation tasks, \eg, 83.8\% top-1 accuracy on ImageNet, 51.7\% box AP and 44.8\% mask AP on COCO val2017, and 49.9\% mIoU on ADE20K, surpassing previous transformer-based and MLP-based models with better trade-off for accuracy and throughput. Code is available at \href{https://github.com/ggjy/Hire-Wave-MLP.pytorch}{https://github.com/ggjy/Hire-Wave-MLP.pytorch}.

\end{abstract}

\section{Introduction}
Attention mechanism based transformers have shown great superiority in the realm of natural language processing in recent years. Several works such as ViT~\cite{vit} and DeiT~\cite{deit} propose to transfer the transformers into visual recognition tasks~\cite{han2020survey}, and have achieved awesome results which are comparable with conventional convolutional neural networks (CNNs). However, the heavy computational burdens caused by the self-attention modules in transformers withhold the models from better trade-off between accuracy and latency. Recently, models composed of only multi-layer perceptrons (MLPs) have become a new trend in vision community~\cite{mixer,resmlp}. These MLP-based models can achieve comparable results with CNNs while discarding the heavy self-attention module. For example, MLP-Mixer~\cite{mixer} extracts per-location information through MLPs that are applied to every image patch, and captures long-range information through MLPs that are applied across patches. 

Although MLP-Mixer can obtain the global receptive field, there are two intractable flaws that prevent the model from becoming a more general backbone for vision tasks: (i) The number of the patches (tokens) will change as the input size changes, which means it cannot be directly fine-tuned at other resolutions that are different from those used in pre-training phase, making MLP-Mixer infeasible to be transferred into downstream vision tasks such as detection and segmentation. (ii) MLP-Mixer rarely explores the local information, which is demonstrated as an useful inductive bias in both CNNs and transformer-based architectures~\cite{resnet,xiao2021early}. The above challenges naturally motivate us to explore an efficient MLP-based architecture which can encode both local and global information while being compatible with flexible input resolutions at the same time. 

\begin{figure*}[t!]
\centering
\includegraphics[width=\textwidth]{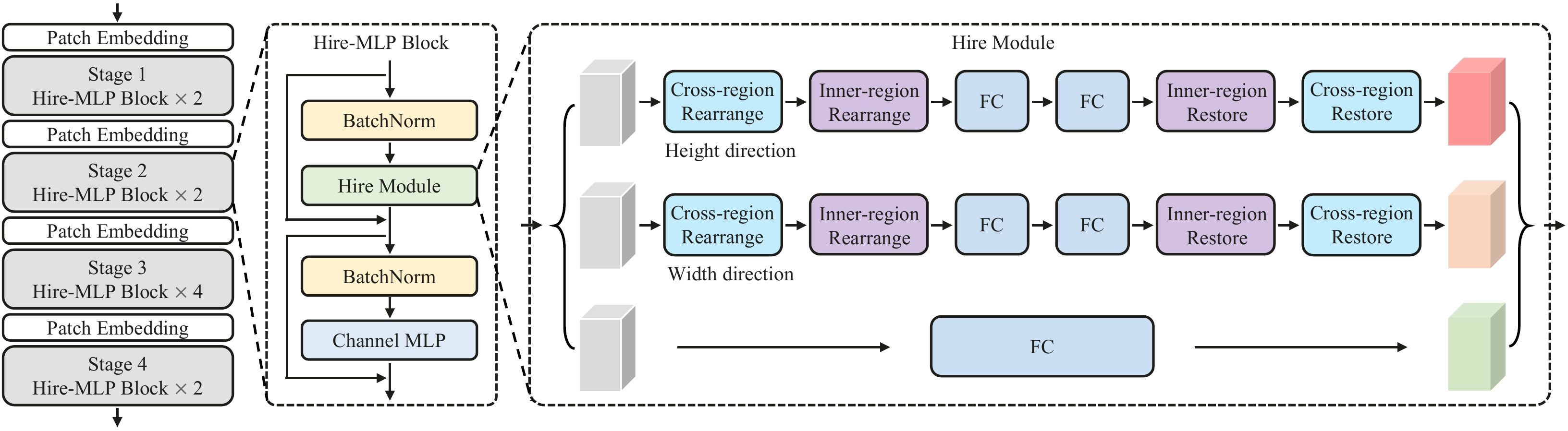}
\vspace{-0.6cm}
\caption{\small The overall architecture of the proposed Hire-MLP-Tiny. More details and other variants of Hire-MLP can be found in Table~\textcolor{red}{A.1} in supplementary materials. Rearrangement layer and restoration layer in hire module are illustrated in Figure~\ref{fig:hire_module}.}
\vspace{-0.2cm}
\label{fig:hire_mlp_arch}
\end{figure*}

\begin{figure}                                                          
\centering       
\small
\includegraphics[width=\columnwidth]{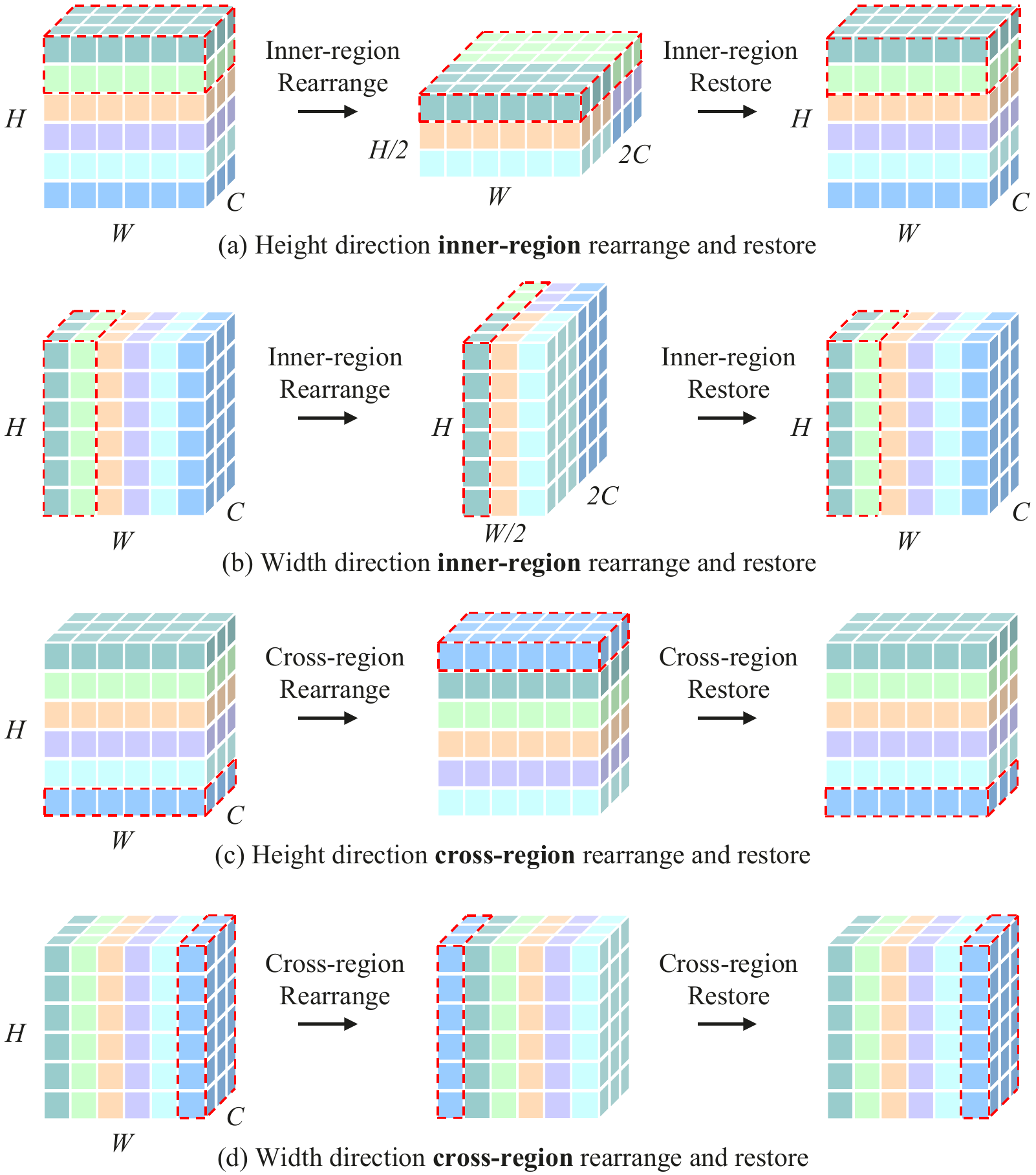}
\vspace{-0.6cm}
\caption{Illustration of the inner-region and cross-region rearrangement operations in hire module.}
\vspace{-0.3cm}
\label{fig:hire_module}
\end{figure}

To address the two aforementioned challenges, we propose the Hire-MLP, which innovates the existing MLP-based models by using hierarchical rearrangement operations. Taking the first challenge into account, the sequence of tokens in MLP-Mixer~\cite{mixer} are denoted as $X\in \R^{HW\times C}$, where $HW$ and $C$ denote the number of tokens and channels, respectively. MLP-Mixer first uses a token-mixing MLP which acts on the columns of $X$ to map $\mathbb{R}^{HW} \mapsto \mathbb{R}^{HW}$, and then uses a channel-mixing MLP which acts on the rows of $X$ to map $\mathbb{R}^{C} \mapsto \mathbb{R}^{C}$. 
The parameters of the token-mixing MLP are configured by the number of tokens $HW$, which depends on the resolution of input images and results in the first challenge.
To this end, we construct our Hire-MLP merely by channel-mixing MLPs applied on the channel dimension.
As for the second challenge, we build the blocks of Hire-MLP based on hierarchical rearrangements and channel-mixing MLPs. The hierarchical rearrangement operation consists of the inner-region rearrangement and the cross-region rearrangement, in which both the local and global information can be easily captured in both height and width directions. We first split the input tokens into multiple regions along the height/width directions, and leverage the inner-region rearrangement operation to shuffle all adjacent tokens belonging to the same region into a one-dimensional vector, followed by two fully connected layers to capture local information within these features. After that, this one-dimensional vector is restored back to the initial arrangement, as illustrated in Figure~\ref{fig:hire_mlp_arch}. For the communication between tokens from different regions, a cross-region rearrangement operation is implemented by shifting all the tokens along a specific direction, as shown in Figure~\ref{fig:hire_module}(c)(d). Such hierarchical rearrangement operation enables our model to obtain both local and global information, and can easily handle the flexible input resolutions.

To be specific, our Hire-MLP has a hierarchical architecture similar to conventional CNNs~\cite{resnet} and recently proposed transformers~\cite{swin,pvt} to generate pyramid feature representations for downstream vision tasks. The overall architecture is shown in Figure~\ref{fig:hire_mlp_arch}. After the first projection layer, the resulting feature $X\in \mathbb{R}^{H\times W\times C}$ is then fed into a sequence of Hire-MLP blocks. Hire module is a key component in Hire-MLP block, which consists of three independent branches. The first two branches consist of a cross-region rearrangement layer, an inner-region rearrangement layer, two channel-mixing fully connected (FC) layers, an inner-region restore layer and a cross-region restore layer to capture local and global information along specific direction, \ie, the height and the width direction. The last branch is built upon a simple channel-mixing FC layer to capture channel information. Compared to existing MLP-based models that spatially shift features in different directions~\cite{asmlp,s2mlp} or leverage a new cycle fully connected operator~\cite{cyclemlp}, our Hire-MLP needs only the channel-mixing MLPs and rearrangement operations. Furthermore, the rearrangement operations can be easily realized by commonly used reshape and padding operations in Pytorch/Tensorflow. And our Hire-MLP is completely capable to serve as a versatile backbone for various computer vision tasks.

Experiments show that Hire-MLP can largely improve the performances of existing MLP-based models on various tasks, including image classification, object detection, instance segmentation, and semantic segmentation. For example, the Hire-MLP-Small attains an 82.1\% top-1 accuracy on ImageNet, outperforming Swin-T~\cite{swin} significantly with a higher throughput. Scaling up the model to larger sizes, we can further obtain 83.2\% and 83.8\% top-1 accuracy. Using Hire-MLP-Small as backbone, Cascade Mask R-CNN achieves 50.7\% box AP and 44.2\% mask AP on COCO val2017. In addition, Hire-MLP-Small obtains 46.1\% single-scale mIoU on ADE20K, which has an improvement of +1.6\% mIoU over Swin-T, demonstrating that Hire-MLP can achieve a better accuracy-latency trade-off than prior MLP-based and transformer-based architectures.

\section{Related Work}
\label{sec:related-work}
\noindent\textbf{CNN-based Models.} LeCun~\etal~proposed the classical LeNet~\cite{lecun1998gradient} in 1990s, which contained most of the basic components of modern CNNs (\eg, convolution and pooling). In ILSVRC 2012 contest, AlexNet~\cite{alexnet} achieved far higher performance than others and drew much attention to CNNs. VGGNet~\cite{vggnet} constructed a plain model by stacking only convolutional layers with kernel size of $3\times3$. GoogLeNet~\cite{googlenet} designed an inception module containing multiple branches to fuse features from diverse receptive fields. To train an extremely deep model for better performance, ResNet~\cite{resnet,he2016identity} skipped multiple layers with an identity projection to alleviate gradient vanishing or exploding. In addition to accuracy, efficiency also plays a crucial part in the practical implementation of CNN-based models, especially on resource-limited devices such as mobile phones. MobileNet~\cite{mobilenet} adopted depth-wise convolutions to aggregate spatial information. ShuffleNet~\cite{shufflenet} introduced the shuffle operation to complement the information loss caused by group convolutions. Such operation can exchange the information across different groups. These well-designed CNNs have been widely used in various tasks such as image recognition~\cite{resnet}, object detection~\cite{ren2015faster}, semantic segmentation~\cite{deeplabv3+} and video analysis~\cite{karpathy2014large}.

\vspace{0.1cm}
\noindent\textbf{Transformer-based Models.} The classical transformer model~\cite{vaswani2017attention} was originally designed to tackle natural language processing (NLP) tasks such as machine translation and English constituency parsing. Recently, Dosovitskiy~\etal~\cite{vit} introduced it to vision community by splitting an image into multiple patches and taking each patch as a token in NLP. Vision transformers can accommodate more training data and achieve higher performance compared to CNNs when the dataset is large enough. Touvron~\etal~\cite{deit} explored how to train data-efficient vision transformers and proposed a new distillation strategy. Extensive works~\cite{pvt,swin,fang2021makes,tnt,t2t,cvt,ceit,twins} were proposed to design the architecture of transformers. For example, PVT~\cite{pvt} designed a pyramid-like structure, where the spatial sizes of feature maps are reduced stage-by-stage, and validated the efficiency of transformers on dense prediction tasks such as object detection and semantic segmentation. TNT~\cite{tnt} embedded small transformer blocks in original modules to capture local information. T2T-ViT~\cite{yuan2021tokens} improved the tokenization process of input images, and proposed a layer-wise Tokens-to-Token transformation module by recursively aggregating neighboring tokens. The information of images can be preserved more sufficiently compared to the simple tokenization with a single layer. Considering the high computational cost of self-attention mechanism, Swin Transformer~\cite{swin} calculated the attention between different tokens in shifted local windows, reducing the computational cost from quadratic to linear complexity. However, the self-attention mechanism is still computational expensive and relatively slow on devices like GPUs.

\vspace{0.1cm}
\noindent\textbf{MLP-based Models.} Considering the large computational cost of attention modules in transformers, simple and efficient models that consist of only multi-layer perceptrons (MLPs) are proposed~\cite{mixer,resmlp}. For example, MLP-Mixer~\cite{mixer} used token-mixing MLP and channel-mixing MLP to capture the relationship between tokens and between channels, respectively. Concurrently, the performance of MLP-based models are further improved by designing new architectures~\cite{vip,cyclemlp,asmlp,s2mlp,s2mlpv2}. CycleMLP~\cite{cyclemlp} introduced a cycle fully connected layer to capture the spatial information, which replaces token-mixing MLP in~\cite{mixer}. AS-MLP~\cite{asmlp} shifted tokens along vertical and horizontal directions to get an axial receptive field. $S^2$-MLP~\cite{s2mlp} also used the shift operation to achieve cross-patch communications. Different from them, our method can simultaneously capture both local and global spatial information by a hierarchical rearrangement operation, \ie, rearranging tokens in/cross local regions, which also achieves a better trade-off between high performance and computational efficiency.

\section{Method}

\subsection{Hire-MLP Block}
The proposed Hire-MLP architecture is constructed by stacking multiple Hire-MLP blocks, as detailed in Figure~\ref{fig:hire_mlp_arch}. Similar to ViT~\cite{vit} and MLP-Mixer~\cite{mixer}, each Hire-MLP block consists of two sub-blocks, \ie, the proposed hire module and channel MLP in~\cite{mixer}, aggregating spatial information and channel information, respectively. Given the input feature $X\in \R^{H\times W \times C}$
with  height $H$, width $W$, and channel number $C$, a Hire-MLP block can be formulated as:
\begin{equation}
	\label{eq-block}
	\begin{aligned}
		&Y= {\rm Hire\text{-}Module} ({\rm BN} (X))+X, \\
		&Z= {\rm Channel\text{-}MLP} ({\rm BN} (Y))+Y,
	\end{aligned}
\end{equation}
where $Y$ and $Z$ are intermediate feature and output feature of the block, respectively. BN denotes the batch normalization~\cite{bn}. The whole Hire-MLP architecture is constructed by iteratively stacking the Hire-MLP block (Eq.~\ref{eq-block}).  
Compared with MLP-Mixer~\cite{mixer}, the major difference is that we replace token-mixing MLP in MLP-Mixer with the proposed hire module and have successfully managed to capture the relationship between different tokens effectively.

\subsection{Hierarchical Rearrangement Module}
\label{sec:hire-module}
In MLP-Mixer~\cite{mixer}, the token-mixing MLP takes linearly flattened tokens as input, and uses fully connected layers to capture the cross-location information. As the dimension of fully connected layers is fixed, it is not compatible with sequences of variable lengths on dense prediction tasks such as object detection and semantic segmentation. Besides, each token-mixing operation captures and aggregates the global information, while some crucial local information might be neglected. In this section, we propose the \textbf{hi}erarchical \textbf{re}arrangement (hire) module to replace the token-mixing MLP in \cite{mixer} and address these challenges accordingly. Briefly, the \emph{inner-region rearrangement} operation in hire module can help capture the local information of tokens in a pre-defined region, while the global information can be captured through \emph{cross-region rearrangement} operation. And credited to the proposed \emph{region partition}, the size of each region remains the same when taking inputs of different sizes. Therefore, our hire module can naturally tackle with sequences of variable lengths and has linear computational complexity with respect to input size. 
In the following, we will introduce the \emph{region partition}, \emph{inner-region rearrangement}, and \emph{cross-region rearrangement} in details.

\vspace{0.1cm}
\noindent\textbf{Region Partition.} We first split the input features into multiple regions, and perform the \emph{inner-region rearrangement} on tokens in each region. The feature can be split along both width and height directions. Taking the height direction \emph{inner-region rearrangement} as an example, the input feature $X$ of shape $H \times W \times C$ will be divided into $g$ regions, \ie, $X=[X_1,X_2,\cdots,X_g]$. Each region $X_i \in \R^{h \times W \times C}$ contains $h$ tokens along the height direction, where $h=H/g$.

\vspace{0.1cm}
\noindent\textbf{Inner-region Rearrangement.} Given an input feature $X_i \in\R^{h \times W \times C}$ of the $i$-th region along the height direction, different tokens will exchange the information adequately through \emph{inner-region rearrangement} operation. Specifically, we concatenate all tokens in $X_i$ along the channel dimension, and get the rearranged feature $X^c_i$ with the shape of $W \times hC$~($h=2$ in Figure~\ref{fig:hire_module}(a)). Then $X^c_i$ is sent to an MLP module $\F$ to mix information along the last dimension and produce output feature $X^o_i \in\R^{W \times hC}$. For efficiency, the MLP $\F$ is implemented by two linear projections with bottleneck, \ie, the feature is first reduced to $W \times \frac{C}{2}$ and then restored to $W \times hC$. A non-linear activation function (\eg, ReLU~\cite{relu} and GeLU~\cite{gelu}) and normalization layer (\eg, BN~\cite{bn} and LN~\cite{ln}) can also be inserted into linear projections to enhance representation ability and stabilize training. At last, the output feature  $X^o_i \in\R^{W \times hC}$ is restored to the original shape for next module, \ie, it is split into multiple tokens along the last dimension to get feature $X'_i \in\R^{h \times W \times C}$. In this way, different tokens in each region can be mixed adequately for generating output features.

\vspace{0.1cm}
\noindent\textbf{Cross-region Rearrangement.} Although the \emph{inner-region rearrangement} enables the communication among tokens in a local region, the receptive field of output feature is limited by the size of each region. Here we introduce the \emph{cross-region rearrangement} operation that exchanges information across different regions by shifting tokens along the height/width direction, and in return enables the model to aggregate global spatial information.

The \emph{cross-region rearrangement} is implemented by recurrently shifting all the tokens along a specific direction with a given step size $s$, as illustrated in Figure~\ref{fig:hire_module}(c) ($s=1$ along the height direction) and Figure~\ref{fig:hire_module}(d) ($s=1$ along the width direction). After shifting, tokens included in the local region split by \emph{region partition} will change. It is worth noting that this operation can be easily accomplished by the ``circular padding" in Pytorch/Tensorflow. To get a global receptive field, the \emph{cross-region rearrangement} operations are inserted before the \emph{inner-region rearrangement} operation every two blocks. The positions of shifted tokens are also restored after the \emph{inner-region restoration} operation to preserve the relative position between different tokens. And this restoration can further boost the accuracy of our Hire-MLP, as shown in Table~\ref{table:ablation-component}.

Note that Zhang~\etal~\cite{shufflenet} uses the channel shuffle operation to communicate across different groups, which disorganizes channels totally. In the contrast, our proposed \emph{cross-region rearrangement} preserves the relative position between different tokens. We argue that the relative position is vital to achieve high representation ability, and related ablation study for these two strategies is investigated in Table~\ref{table:ablation-communication}. We also visualize the feature maps after two cross-region rearrangement manners (ShuffleNet~\cite{shufflenet} manner vs. our shifted manner) in supplementary materials. 

\vspace{0.1cm}
\noindent\textbf{Hire Module.} Considering an input feature $X$ of size $H\times W\times C$, the spatial information communication is conducted within two branches, \ie, along the height direction and the width direction. Inspired by shortcut connections in ResNet~\cite{resnet} and ViP~\cite{vip}, an extra branch without spatial communication is also added, where only a fully connected layer is leveraged to encode information along the channel dimension. The input $X$ is sent to above three branches to get features $X'_W$, $X'_H$, and $X'_C$, respectively. Then the output feature $X'$ is obtained by summing up these features, \ie, $X'=X'_W+X'_H+X'_C$, as depicted in Figure~\ref{fig:hire_mlp_arch}.

\vspace{0.1cm}
\noindent\textbf{Complexity Analysis.} In hire module, the fully connected layer (FC) consumes the major memory and computational cost. Consider the height direction branch in Figure~\ref{fig:hire_mlp_arch}, given an input feature $X\in\R^{H\times W \times C}$, we first split it into $H/h$ regions with the shape of $h\times W \times C$. And the shape of the feature after \emph{inner-region rearrangement} is $H/h\times W\times hC$. We empirically set the channel dimension in bottleneck to $C/2$, thereby this branch occupies $hC\times\frac{C}{2}\times2=hC^2$~parameters and $\frac{H}{h}\times W\times hC\times\frac{C}{2}\times2=HWC^2$~FLOPs. Considering a hire module with three branches (height, width, and channel), the total parameters and FLOPs are $(2hC^2+C^2)$ and $3HWC^2$, respectively. 

\subsection{Overall Architecture}
An overview of the Hire-MLP-Tiny architecture is shown in Figure~\ref{fig:hire_mlp_arch}, more details and other variants of Hire-MLP are presented in Table~\textcolor{red}{A.1} in supplementary materials. We adopt a pyramid-like architecture for Hire-MLP following the commonly used design of CNNs~\cite{resnet,vggnet} and vision transformers~\cite{pvt,swin}. It first splits the input image into patches (tokens) by a patch embedding layer~\cite{pvtv2}. Then two Hire-MLP blocks referred to as ``Stage 1" are applied on the tokens above. As the network gets deeper, the number of tokens is reduced by another patch embedding layer and output channels are doubled at the same time. Especially, the whole architecture contains four stages, where the feature resolution reduces from $\frac{H}{4} \times \frac{W}{4}$ to $\frac{H}{32} \times \frac{W}{32}$ and the output dimension increases accordingly. The pyramid architecture aggregates the spatial feature for extracting semantic information, which can be applied to image classification, object detection, and semantic segmentation.

\footnotetext[1]{AS-MLP~\cite{asmlp} reported the throughput under the mixed precision training mode, here we reproduce it and report the throughput under the pure precision training mode for a fair comparison with other methods.}

We develop diverse variants of Hire-MLP architectures with different memory and computational cost. The ``Base" model~(Hire-MLP-Base) contains \{4, 6, 24, 3\} layers for each stage. ``Tiny" and ``Small" variants have fewer layers to realize efficient implementation, while the ``Large" variant has larger representation capacity to achieve higher performance. Detailed configurations can also be found in  supplementary materials.

\section{Experiments}
In this section, we investigate the effectiveness of Hire-MLP architectures by conducting experiments on several vision tasks. We first compare the proposed Hire-MLP with previous state-of-the-art models for image classification on ImageNet-1K~\cite{imagenet}, and then we ablate the important design elements of Hire-MLP. We also present the results of object detection and semantic segmentation on COCO~\cite{coco} and ADE20K~\cite{ade20k}, respectively.

\subsection{Image Classification on ImageNet}

\begin{table}[!t]
\renewcommand{\arraystretch}{0.88}
\setlength\tabcolsep{3.8pt}
\small
\centering
\begin{tabular}{l|c|c|c|c}
\toprule
Network & Params & FLOPs & \tabincell{c}{Throughput \\ (image / s)} & Top-1 \tabularnewline
\midrule
\multicolumn{5}{c}{CNN-based} \\
\midrule		
RegNetY-4GF~\cite{regnety} & 39M & 4.0G & 1156.7 & 81.0 \tabularnewline
RegNetY-16GF~\cite{regnety} & 84M & 16.0G & 334.7 & 82.9 \tabularnewline
EfficientNet-B4~\cite{efficientnet} & 19M & 4.2G & 349.4 & 82.9 \tabularnewline 
EfficientNet-B5~\cite{efficientnet} & 30M & 9.9G & 169.1 & 83.6 \tabularnewline 
EfficientNet-B6~\cite{efficientnet} & 43M & 19.0G & 96.9 & 84.0 \tabularnewline 
\midrule
\multicolumn{5}{c}{Transformer-based}  \tabularnewline
\midrule
DeiT-S~\cite{deit} & 22M & 4.6G & 940.4 & 79.8 \tabularnewline
T2T-ViT$_\mathrm{t}$-14~\cite{t2t} & 22M & 5.2G & - & 80.7 \tabularnewline
Swin-T~\cite{swin} & 29M & 4.5G & 755.2 & 81.3 \tabularnewline
CPVT-S-GAP~\cite{cpvt} & 22M & 4.6G & 942.3 & 81.5 \tabularnewline
PVT-M~\cite{pvt} & 44M & 6.7G & 528.1 & 81.2 \tabularnewline
PVT-L~\cite{pvt} & 61M & 9.8G & 358.8 & 81.7 \tabularnewline
T2T-ViT$_\mathrm{t}$-24~\cite{t2t} & 64M & 15.0G & - & 82.6 \tabularnewline
TNT-B~\cite{tnt} & 66M & 14.1G & - & 82.9 \tabularnewline 
Swin-B~\cite{swin} & 88M & 15.4G & 278.1 & 83.5 \tabularnewline
\midrule
\multicolumn{5}{c}{MLP-based}  \tabularnewline
\midrule
gMLP-Ti~\cite{gmlp} & 6M & 1.4G & - & 72.3 \tabularnewline
CycleMLP-B1~\cite{cyclemlp} & 15M & 2.1G & 1038.4\rlap{$^\ddagger$} & 78.9 \tabularnewline
Hire-MLP-Tiny~\textbf{(ours)} & 18M & 2.1G & 1561.7 & \textbf{79.7} \tabularnewline 
\midrule
ResMLP-S12~\cite{resmlp} & 15M & 3.0G & 1415.1 & 76.6 \tabularnewline
ViP-Small/7~\cite{vip} & 25M & - & 719.0 & 81.5 \tabularnewline
AS-MLP-T$^*$~\cite{asmlp} & 28M & 4.4G & 863.6\rlap{$^\ddagger$} & 81.3 \tabularnewline
CycleMLP-B2~\cite{cyclemlp} & 27M & 3.9G & 640.6\rlap{$^\ddagger$} & 81.6 \tabularnewline
Hire-MLP-Small~\textbf{(ours)} & 33M & 4.2G & 807.6 & \textbf{82.1} \tabularnewline 
\midrule
Mixer-B/16~\cite{mixer} & 59M & 12.7G & - & 76.4 \tabularnewline
S$^2$-MLP-deep~\cite{s2mlp} & 51M & 10.5G & - & 80.7 \tabularnewline 
ResMLP-B24~\cite{resmlp} & 116M & 23.0G & 231.3 & 81.0 \tabularnewline
ViP-Medium/7 \cite{vip} & 55M & - & 418.0 & 82.7 \tabularnewline
CycleMLP-B4 \cite{cyclemlp} & 52M & 10.1G & 320.8\rlap{$^\ddagger$} & 83.0 \tabularnewline
AS-MLP-S$^*$~\cite{asmlp} & 50M & 8.5G & 478.4\rlap{$^\ddagger$} & 83.1 \tabularnewline
Hire-MLP-Base~\textbf{(ours)} & 58M & 8.1G & 440.6 & \textbf{83.2} \tabularnewline 
\midrule	
S$^2$-MLP-wide~\cite{s2mlp} & 71M & 14.0G & - & 80.0 \tabularnewline 
CycleMLP-B5~\cite{cyclemlp} & 76M & 12.3G & 246.9\rlap{$^\ddagger$} & 83.2 \tabularnewline
gMLP-B~\cite{gmlp} & 73M & 15.8G & - & 81.6 \tabularnewline
ViP-Large/7~\cite{vip} & 88M & - & 298.0 & 83.2 \tabularnewline
AS-MLP-B$^*$~\cite{asmlp} & 88M & 15.2G & 312.4\rlap{$^\ddagger$} & 83.3 \tabularnewline
Hire-MLP-Large~\textbf{(ours)} & 96M & 13.4G & 290.1 & \textbf{83.8} \tabularnewline 
\bottomrule
\end{tabular}
\vspace{-0.3cm}
\caption{Experimental results of different networks on ImageNet-1K. Throughput is measured as the number of images that we can process per second on a single V100 GPU following~\cite{deit,swin}. $^*$~means AS-MLP~\cite{asmlp} accelerates the AS operation by CUDA implementation. $^\ddagger$ means the throughput result is reproduced by us\protect\footnotemark.}
\vspace{-0.2cm}
\label{table:imagenet-1k}
\end{table}

\begin{table}[h]
\renewcommand{\arraystretch}{0.9}
\setlength\tabcolsep{4pt}
\small
\centering
\begin{tabular}{c|c|c|c}
\toprule
Num. of $h$ and $w$ & Top-1 (\%) & Num. of $h$ and $w$ & Top-1 (\%) \tabularnewline \midrule
(2, 2, 2, 2) & 81.62 & (2, 2, 3, 3) & 81.73 \tabularnewline \midrule
(3, 2, 2, 2) & 81.82 & (3, 3, 2, 2) & 81.78 \tabularnewline \midrule
(3, 3, 3, 2) & 81.87 & (3, 3, 3, 3) & 81.79 \tabularnewline \midrule
(4, 3, 3, 2) & \textbf{82.07} & (4, 3, 3, 3) & 81.86 \tabularnewline \midrule
(4, 4, 3, 3) & 81.81 & (4, 4, 4, 4) & 81.72 \tabularnewline \midrule
(5, 4, 3, 3) & 81.74 & (6, 4, 3, 3) & 81.49 \tabularnewline
\bottomrule
\end{tabular}
\vspace{-0.3cm}
\caption{Ablation study on the number of tokens in each region in \emph{Region Partition}. Given an input feature of size $H$$\times$$W$$\times$$C$, we split it into $H/h$ ($W/w$) regions along the height (width) direction, and the size of each region is $h$$\times$$W$$\times$$C$. We set $h$$=$$w$ as default for 224$\times$224 input resolution. For example, (4, 3, 3, 2) indicates $h$ and $w$ are set to 4, 3, 3, and 2 for stage 1, stage 2, stage 3, and stage 4, respectively. The step size $s$ here is set to (2, 2, 1, 1).}
\label{table:ablation-hw}
\end{table}

\begin{table}[h]
\renewcommand{\arraystretch}{0.9}
\small
\centering
\begin{tabular}{c|c|c|c}
\toprule
Num. of $s$ & Top-1 (\%) & Num. of $s$ & Top-1 (\%) \tabularnewline \midrule
(0, 0, 0, 0) & 81.18 & (1, 1, 1, 1) & 81.88 \tabularnewline \midrule
(2, 2, 1, 1) & \textbf{82.07} & (2, 2, 2, 2) & 81.71 \tabularnewline
\bottomrule
\end{tabular}
\vspace{-0.3cm}
\caption{Ablation study about the step size of shifted tokens ($s$) in \emph{cross-region rearrangement}. For example, (2, 2, 1, 1) means $s$ is set to 2, 2, 1, and 1 for stage 1, stage 2, stage 3, and stage 4, respectively. (0, 0, 0, 0) indicates there is no \emph{cross-region rearrangement} in Hire-MLP. The $h$ and $w$ here are set to (4, 3, 3, 2).}
\label{table:ablation-shift}
\end{table}

\begin{table}[h]
\renewcommand{\arraystretch}{0.9}
\setlength\tabcolsep{4pt}
\small
\centering
\begin{tabular}{c|c|c|c}
\toprule
Padding mode & Top-1 (\%) & Padding mode & Top-1 (\%) \tabularnewline \midrule
Zero padding & 81.62 & Circular padding & \textbf{82.07} \tabularnewline \midrule
Reflect padding & 81.48 & Replicated padding & 81.60 \tabularnewline
\bottomrule
\end{tabular}
\vspace{-0.3cm}
\caption{Different padding modes for \emph{inner-region rearrangement}.}
\label{table:ablation-padding}
\end{table}

\begin{table}[t]
\renewcommand{\arraystretch}{0.9}
\small
\centering
\begin{tabular}{c|c}
\toprule
Model & Top-1 (\%) \tabularnewline \midrule
Hire-MLP-Small & \textbf{82.07} \tabularnewline \midrule
w/o cross-region restore & 81.70 \tabularnewline \midrule
w/o cross-region rearrange and restore & 81.18 \tabularnewline \midrule
w/o inner-region rearrange and restore & 80.17 \tabularnewline \midrule
w/o extra FC branch & 81.32 \tabularnewline
\bottomrule
\end{tabular}
\vspace{-0.3cm}
\caption{Impact of different components in hire module.}
\label{table:ablation-component}
\end{table}

\begin{table}[t]
\renewcommand{\arraystretch}{0.9}
\small
\centering
\begin{tabular}{c|c}
\toprule
Model & Top-1 (\%) \tabularnewline \midrule
Shifted manner & \textbf{82.07} \tabularnewline \midrule
ShuffleNet manner~\cite{shufflenet} & 80.90 \tabularnewline
\bottomrule
\end{tabular}
\vspace{-0.3cm}
\caption{Different manners for cross-region communication.}
\vspace{-0.2cm}
\label{table:ablation-communication}
\end{table}

\begin{table}[t]
\renewcommand{\arraystretch}{0.9}
\small
\centering
\begin{tabular}{c|c|c|c}
\toprule
Num. of FC layer & \# Params & \# FLOPs & Top-1 (\%) \tabularnewline \midrule
1 & 49.65M & 5.65G & 82.15 \tabularnewline \midrule
2 & 33.11M & 4.24G & 82.07 \tabularnewline \midrule
3 & 32.98M & 4.23G & 81.81 \tabularnewline \midrule
4 & 33.26M & 4.24G & 81.85 \tabularnewline
\bottomrule
\end{tabular}
\vspace{-0.3cm}
\caption{Ablation study about the number of intermediate FC layers in first two branches in hire module.}
\vspace{-0.2cm}
\label{table:ablation-fc}
\end{table}

\noindent\textbf{Experimental Settings.} We conduct experiments on the challenging ImageNet-1K~\cite{imagenet}, which is a image classification benchmark containing 1.28M training images and 50K validation images of 1000 classes. ImageNet-1K is also utilized to conduct the ablation studies. For fair comparisons with recent works, we adopt the same training and augmentation strategy as those in DeiT~\cite{deit}, \ie, models are trained for 300 epochs using the AdamW~\cite{adamw} optimizer with weight decay 0.05 and the batch size of 1024. We use a linear warmup for early 20 epochs, the initial learning rate is set to 1e-3 and gradually drops to 1e-5. The data augmentation methods include Rand-Augment~\cite{randaugment}, MixUp~\cite{mixup}, CutMix~\cite{cutmix}, Label Smoothing~\cite{labelsmooth}, Random Erasing~\cite{randomerasing}, and DropPath~\cite{droppath}. All models are trained on 8 NVIDIA Tesla V100 GPUs, we report the experimental results with single-crop top-1 accuracy, parameters, FLOPs and throughput.

\vspace{0.1cm}
\noindent\textbf{Main Results.} We compare the proposed Hire-MLP with previous CNN-based, transformer-based, and MLP-based models on Imagenet as shown in Table~\ref{table:imagenet-1k}. The resolution of input image is set to 224 $\times$ 224. For example, our Hire-MLP-Small achieves 82.1\% top-1 accuracy with only 4.2G FLOPs, which is better than all other existing MLP-based models. When compared to recently proposed AS-MLP~\cite{asmlp} and CycleMLP~\cite{cyclemlp}, our Hire-MLP can obtain better performances (+0.5$\sim$0.8) without any complicated shift operations or variants of fully connected layer. Scaling up our model to 8.1G and 13.1G can achieve 83.2\% and 83.8\% top-1 accuracy, respectively. The superiority of Hire-MLP demonstrates that the proposed hire module can better capture both local and global information, which is crucial for classification. In addition, we show the comparison with conventional CNN-based and transformer-based models. When compared to transformer-based models such as DeiT~\cite{deit}, Swin Transformer~\cite{swin}, and PVT~\cite{tnt}, our model can get better results with a faster inference speed. When compared to CNN-based architectures such as RegNetY~\cite{regnety}, our Hire-MLP can achieve better results with smaller model size and lower computational cost. However, there is still a small gap between our model and the state-of-the-art EfficientNet-B6. We argue that MLP-based architectures have their unique advantages of simplicity and faster inference speed (290.1 vs. 96.9), and there are still opportunities for further enhancements for MLP-based models.

\begin{table*}[t]
\renewcommand{\arraystretch}{0.88}
\setlength\tabcolsep{3.4pt}
\begin{tabular}{l|c|c|ccc|c|ccc|ccc}
\toprule
\multirow{2}{*}{Backbone} &\multicolumn{5}{c|}{RetinaNet 1$\times$} &\multicolumn{7}{c}{Mask R-CNN 1$\times$} \\
\cline{2-13} 
& Param / FLOPs & AP & AP$_\mathrm{S}$ &AP$_\mathrm{M}$ & AP$_\mathrm{L}$ & Param / FLOPs & AP$^{\rm b}$ & AP$_{50}^{\rm b}$ &AP$_{75}^{\rm b}$  &AP$^{\rm m}$ &AP$_{50}^{\rm m}$ & AP$_{75}^{\rm m}$\\
\midrule

ResNet18~\cite{resnet}  & 21.3M / 188.7G & 31.8 & 16.3 & 34.3 & 43.2 & 31.2M / 207.3G & 34.0 & 54.0 & 36.7 & 31.2 & 51.0 & 32.7 \\
PVT-Tiny~\cite{pvt} & 23.0M / 189.5G & 36.7 & 22.6 & 38.8 & 50.0 & 32.9M / 208.1G & 36.7 & 59.2 & 39.3 & 35.1 & 56.7 & 37.3 \\
CycleMLP-B1~\cite{cyclemlp} & 24.9M / 195.0G & 38.6 & 21.9 & 41.8 & \textbf{50.7} & 34.8M / 213.6G & 39.4 & 61.4 & 43.0 & 36.8 & 58.6 & 39.1 \\
Hire-MLP-Tiny & 27.8M / 195.3G & \textbf{38.9} & \textbf{24.9} & \textbf{42.7} & \textbf{50.7} & 37.7M / 213.8G & \textbf{39.6} & \textbf{61.7} & \textbf{43.1} & \textbf{37.0} & \textbf{59.1} & \textbf{39.6} \\
\midrule

ResNet50~\cite{resnet} & 37.7M / 239.3G & 36.3 & 19.3 & 40.0 & 48.8 & 44.2M / 260.1G & 38.0 & 58.6 & 41.4 & 34.4 & 55.1 & 36.7 \\
PVT-Small~\cite{pvt} & 34.2M / 226.5G & 40.4 & 25.0 & 42.9 & 55.7 & 44.1M / 245.1G & 40.4 & 62.9 & 43.8 & 37.8 & 60.1 & 40.3 \\
CycleMLP-B2~\cite{cyclemlp} & 36.5M / 230.9G & 40.9 & 23.4 & 44.7 & 53.4 & 46.5M / 249.5G & 41.7 & 63.6 & 45.8 & 38.2 & 60.4 & 41.0 \\
Swin-T~\cite{swin} & 38.5M / 244.8G & 41.5 & 25.1 & 44.9 & \textbf{55.5} & 47.8M / 264.0G & 42.2 & 64.6 & 46.2 & 39.1 & 61.6 & 42.0 \\
Hire-MLP-Small & 42.8M / 237.6G & \textbf{41.7} & \textbf{25.3} & \textbf{45.4} & 54.6 & 52.7M / 256.2G & \textbf{42.8} & \textbf{65.0} & \textbf{46.7} & \textbf{39.3} & \textbf{62.0} & \textbf{42.1} \\
\midrule

ResNet101~\cite{resnet} & 56.7M / 315.4G & 38.5 & 21.4 & 42.6 & 51.1 & 63.2M / 336.4G & 40.4 & 61.1 & 44.2 & 36.4 & 57.7 & 38.8 \\
PVT-Medium~\cite{pvt} & 53.9M / 283.1G & 41.9 & 25.0 & 44.9 & 57.6 & 63.9M / 301.7G & 42.0 & 64.4 & 45.6 & 39.0 & 61.6 & 42.1 \\
CycleMLP-B3~\cite{cyclemlp} & 48.1M / 291.3G & 42.5 & 25.2 & 45.5 & 56.2 & 58.0M / 309.9G & 43.4 & 65.0 & 47.7 & 39.5 & 62.0 & 42.4 \\
CycleMLP-B4~\cite{cyclemlp} & 61.5M / 356.6G & 43.2 & 26.6 & 46.5 & 57.4 & 71.5M / 375.2G & 44.1 & 65.7 & 48.1 & 40.2 & 62.7 & 43.5 \\
Swin-S~\cite{swin} & 59.8M / 334.8G & \textbf{44.5} & 27.4 & 48.0 & \textbf{59.9} & 69.1M / 353.8G & 44.8 & 66.6 & 48.9 & 40.9 & 63.4 & \textbf{44.2} \\
Hire-MLP-Base & 68.0M / 316.5G & 44.3 & \textbf{28.0} & \textbf{48.4} & 58.0 & 77.8M / 334.9G & \textbf{45.2} & \textbf{66.9} & \textbf{49.3} & \textbf{41.0} & \textbf{64.0} & \textbf{44.2} \\
\midrule

PVT-Large~\cite{pvt} & 71.1M / 345.7G & 42.6 & 25.8 & 46.0 & 58.4 & 81.0M / 364.3G & 42.9 & 65.0 & 46.6 & 39.5 & 61.9 & 42.5 \\
CycleMLP-B5~\cite{cyclemlp} & 85.9M / 402.2G & 42.7 & 24.1 & 46.3 & 57.4 & 95.3M / 421.1G & 44.1 & 65.5 & 48.4 & 40.1 & 62.8 & 43.0 \\
Hire-MLP-Large & 105.8M / 424.5G & \textbf{44.9} & \textbf{28.9} & \textbf{48.9} & \textbf{57.5} & 115.2M / 443.5G & \textbf{45.9} & \textbf{67.2} & \textbf{50.4} & \textbf{41.7} & \textbf{64.7} & \textbf{45.3} \\
\bottomrule
\end{tabular}
\vspace{-0.3cm}
\caption{Object detection and instance segmentation results on COCO val2017. We compare Hire-MLP with other backbones based on RetinaNet and Mask R-CNN frameworks, all models are trained in ``1x" schedule. FLOPs is calculated on 1280$\times$800 input.}
\label{table:coco-1x}
\end{table*}

\begin{table*}[t]
\renewcommand{\arraystretch}{0.88}
\setlength\tabcolsep{4.2pt}
\begin{tabular}{l|c|ccc|ccc|c|ccc|ccc}
\toprule
\multirow{2}{*}{Backbone} &\multicolumn{7}{c|}{Mask R-CNN 3$\times$} &\multicolumn{7}{c}{Cascade Mask R-CNN 3$\times$} \\
\cline{2-15} 
& FLOPs & AP$^{\rm b}$ & AP$_{50}^{\rm b}$ & AP$_{75}^{\rm b}$ & AP$^{\rm m}$ & AP$_{50}^{\rm m}$ & AP$_{75}^{\rm m}$ & FLOPs & AP$^{\rm b}$ & AP$_{50}^{\rm b}$ &AP$_{75}^{\rm b}$  &AP$^{\rm m}$ &AP$_{50}^{\rm m}$ & AP$_{75}^{\rm m}$\\
\midrule  

ResNet50~\cite{resnet} & 260.1G & 41.0 & 61.7 & 44.9 & 37.1 & 58.4 & 40.1 & 738.7G & 46.3 & 64.3 & 50.5 & 40.1 & 61.7 & 43.4 \\
AS-MLP-T~\cite{asmlp} & 260.1G & 46.0 & 67.5 & 50.7 & 41.5 & 64.6 & 44.5 & 739.0G & 50.1 & 68.8 & 54.3 & 43.5 & 66.3 & 46.9 \\
Swin-T~\cite{swin} & 264.0G & 46.0 & \textbf{68.2} & 50.2 & 41.6 & 65.1 & 44.8 & 742.4G & 50.5 & 69.3 & 54.9 & 43.7 & 66.6 & 47.1 \\
Hire-MLP-Small & 256.2G & \textbf{46.2} & \textbf{68.2} & \textbf{50.9} & \textbf{42.0} & \textbf{65.6} & \textbf{45.3} & 734.6G & \textbf{50.7} & \textbf{69.4} & \textbf{55.1} & \textbf{44.2} & \textbf{66.9} & \textbf{48.1} \\
\midrule

Swin-S~\cite{swin} & 353.8G & \textbf{48.5} & \textbf{70.2} & \textbf{53.5} & \textbf{43.3} & \textbf{67.3} & 46.6 & 832.4G & \textbf{51.8} & \textbf{70.4} & \textbf{56.3} & 44.7 & \textbf{67.9} & \textbf{48.5} \\
AS-MLP-S~\cite{asmlp} & 346.0G & 47.8 & 68.9 & 52.5 & 42.9 & 66.4 & 46.3 & 823.8G & 51.1 & 69.8 & 55.6 & 44.2 & 67.3 & 48.1 \\
Hire-MLP-Base & 334.9G & 48.1 & 69.6 & 52.7 & 43.1 & 66.8 & \textbf{46.7} & 813.2G & 51.7 & 70.2 & 56.1 & \textbf{44.8} & 67.8 & \textbf{48.5} \\
\bottomrule
\end{tabular}
\vspace{-0.3cm}
\caption{Instance segmentation results on COCO val2017. Mask R-CNN and Cascade Mask R-CNN are trained in ``3x" schedule.}
\vspace{-0.2cm}
\label{table:coco-3x}
\end{table*}

\begin{table*}[t]
\centering
\renewcommand{\arraystretch}{0.88}
\setlength\tabcolsep{4.4pt}
\begin{tabular}{l|c|c|c|c||l|c|c|c|c|c}
\toprule
\multicolumn{5}{c||}{Semantic FPN}  & \multicolumn{6}{c}{UperNet} \\
\midrule  
Backbone & Param & FLOPs & FPS & SS mIoU & Backbone & Param & FLOPs & FPS & SS mIoU & MS mIoU \\
\midrule  

PVT-Small~\cite{pvt} & 28M & 163G & 43.9\rlap{$^\ddagger$}\, & 39.8 & Swin-T~\cite{swin} & 60M & 945G & 18.5\, & 44.5 & 46.1 \\
CycleMLP-B2~\cite{cyclemlp} & 31M & 167G & 44.5\rlap{$^\ddagger$}\, & 42.4 & AS-MLP-T~\cite{asmlp} & 60M & 937G & 17.7\rlap{$^\ddagger$}\, & - & 46.5 \\
Hire-MLP-Small & 37M & 174G & 47.3\, & \textbf{44.3} & Hire-MLP-Small & 63M & 930G & 19.3\, & \textbf{46.1} & \textbf{47.1} \\
\midrule
CycleMLP-B3~\cite{cyclemlp} & 42M & 229G & 31.0\rlap{$^\ddagger$}\, & 44.5 & ResNet-101~\cite{resnet} & 86M & 1029G & 20.1 & 43.8 & 44.9 \\
GFNet-Base~\cite{gfnet} & 75M & 261G & -\, & 44.8 & Swin-S~\cite{swin} & 81M & 1038G & 15.2\, & 47.6 & 49.5 \\
CycleMLP-B4~\cite{cyclemlp} & 56M & 296G & 23.6\rlap{$^\ddagger$}\, & 45.1 & AS-MLP-S~\cite{asmlp} & 81M & 1024G & 14.4\rlap{$^\ddagger$}\, & - & 49.2 \\
Hire-MLP-Base & 62M & 255G & 31.8\, & \textbf{46.2} & Hire-MLP-Base & 88M & 1011G & 16.0\, & \textbf{48.3} & \textbf{49.6} \\
\midrule

Swin-B~\cite{cyclemlp} & 53M & 274G & 23.4\rlap{$^\ddagger$}\, & 45.2\rlap{$^\dagger$} & Swin-B~\cite{swin} & 121M & 1188G & 13.3\rlap{$^\ddagger$}\, & 48.1 & 49.7 \\
CycleMLP-B5~\cite{cyclemlp} & 79M & 343G & 22.9\rlap{$^\ddagger$}\, & 45.6 & AS-MLP-B~\cite{asmlp} & 121M & 1166G & 11.0\rlap{$^\ddagger$}\, & - & 49.5 \\
Hire-MLP-Large & 99M & 366G & 24.5\, & \textbf{46.6} & Hire-MLP-Large & 127M & 1125G & 13.7\, & \textbf{48.8} & \textbf{49.9} \\

\bottomrule
\end{tabular}
\vspace{-0.3cm}
\caption{Results of semantic segmentation on ADE20K validation set. FLOPs is calculated with the input size of 2048$\times$512. FPS is measured by using a 32G Tesla V100 GPU. $^\dagger$ indicates the results are from GFNet~\cite{gfnet}. $^\ddagger$ indicates the results are measured by us.}
\vspace{-0.2cm}
\label{table:seg}
\end{table*}

\subsection{Ablation Study}
\label{sec:ablation}
The core component in Hire-MLP is the hierarchical rearrangement module (Sec.~\ref{sec:hire-module}). We conduct the ablation studies about the number of tokens in each region in region partition, the number of shifted regions and different rearrangement manners for cross-region rearrangement, the padding mode in inner-region rearrangement, and the number of FC layers in hire module. All ablation experiments are conducted based on the Hire-MLP-Small.

\vspace{0.1cm}
\noindent\textbf{The number of tokens in each region in region partition.}
Table~\ref{table:ablation-hw} investigates how region partition affects the final performance based on Hire-MLP-Small, where $h$ and $w$ denote the size of each region. Consider that the resolution of input image is 224 $\times$ 224 in ImageNet, we set $h=w$ if not specified. A small region size implies few adjacent tokens are mixed via the inner-region rearrangement operation, which emphasizes more on local information. We empirically find that a larger region size is required in lower layer to tackle the feature maps with more tokens and obtain larger receptive fields. When the region size is further increased, the performance will drop slightly. We conjecture that there might be some information loss in the bottleneck structure with the increasing region size.

\vspace{0.1cm}
\noindent\textbf{The step size $s$ of shifted token in cross-region rearrangement.}
The cross-region rearrangement is implemented by shifting tokens with a given step size $s$, whose impact is investigated in Table~\ref{table:ablation-shift}. When the tokens are not shifted, \ie, $s=(0, 0, 0, 0)$, there is no communication between different regions (without cross-region rearrangement operation). Obviously, the lack of global information leads to a bad performance.

\vspace{0.1cm}
\noindent\textbf{The impacts of different padding methods.}
The resolution of the input image from ImageNet~\cite{imagenet} is of size $224\times224$, therefore the shape of output feature in stage~4 is $7\times7$, which is not divisible by any $h$ and $w$. In consequence, we need to pad the feature map. Table~\ref{table:ablation-padding} evaluates the influence of different padding methods. And we find that the ``Circular padding" is the most suitable for the design of hire module.

\vspace{0.1cm}
\noindent\textbf{The impacts of different components in hire module.}
Table~\ref{table:ablation-component} ablates the impacts of different components in hire module (Sec.~\ref{sec:hire-module}). We can find that the inner-region rearrangement is the most important component to capture local information. The cross-region restoration operation can bring about 0.3\% improvement on top-1 accuracy. If we discard the cross-region rearrangement (including the restoration), the model cannot exchange information across different regions, and the performance would drop to 81.18\%. And removing the third branch in Figure~\ref{fig:hire_mlp_arch} would harm the top-1 accuracy by 0.7\%.

\vspace{0.1cm}
\noindent\textbf{Different strategies for cross-region communication.}
We compare two different strategies for cross-region communication in Table~\ref{table:ablation-communication}. The shifted manner achieves better result compared to ShuffleNet manner, indicating shifted manner can preserve more relative position information for model. More details and corresponding visualization of these two strategies can be found in supplementary materials.

\vspace{0.1cm}
\noindent\textbf{The number of FC layers in hire module.}
The bottleneck design of MLP $\F$ in hire module (Sec.~\ref{sec:hire-module}) can help eliminate the heavy burden of FLOPs brought by the increase in channels. Ablation studies about the number of FC layers are reported in Table~\ref{table:ablation-fc}. Although using one FC layer achieves the best performance, the parameter and FLOPs are larger than other counterparts. A bottleneck with two FC layers can obtain a better trade-off between accuracy and computational cost. Furthermore, adding more FC layers cannot bring more benefits, demonstrating that the improvements come from our hierarchical rearrangement operation rather than the increase in the number of FC layers.

\subsection{Object Detection on COCO}
\noindent\textbf{Experimental Settings.} We conduct the object detection and instance segmentation experiments on COCO 2017 benchmark~\cite{coco}, which contains 118K training images and 5K validation images. Following PVT~\cite{pvt} and Swin Transformer~\cite{swin} , we consider three typical object detection frameworks: RetinaNet~\cite{retinanet}, Mask R-CNN~\cite{maskrcnn} and Cascade Mask R-CNN~\cite{cascadercnn} in mmdetection~\cite{mmdetection}. We utilize the single-scale training and multi-scale training for the ``1x" and ``3x" schedules, respectively. More details are introduced in the supplementary materials.

\vspace{0.1cm}
\noindent\textbf{Results.} We report the results of object detection and instance segmentation under different frameworks and training schedules in Table~\ref{table:coco-1x} and Table~\ref{table:coco-3x}, respectively. As shown in Table~\ref{table:coco-1x}, Hire-MLP based RetinaNet and Mask R-CNN consistently surpasses the CNN-based ResNet~\cite{resnet}, transformer-based PVT~\cite{pvt} and MLP-based CycleMLP~\cite{cyclemlp} under similar FLOPs constraints. Consider RetinaNet~\cite{retinanet} as the basic framework, our Hire-MLPs bring consistent +5.8$\sim$7.1 AP gains over ResNets~\cite{resnet} and bring +0.3$\sim$2.2 AP gains over CycleMLPs~\cite{cyclemlp} with slightly larger model size and FLOPs. The results indicate that Hire-MLP can serve as an excellent backbone for object detection. Furthermore, Hire-MLP based Cascade Mask R-CNN surpasses the AS-MLP counterpart by 0.6$\sim$0.7 in both box AP and mask AP with less FLOPs, as shown in Table~\ref{table:coco-3x}.

\subsection{Semantic Segmentation on ADE20K}

\noindent\textbf{Experimental Settings.} We conduct semantic segmentation experiments on ADE20K benchmark~\cite{ade20k}, which contains 20,210 training images and 2,000 validation images. Following~\cite{pvt,cyclemlp,swin}, we consider two typical frameworks: Semantic FPN~\cite{semanticfpn} and UperNet~\cite{upernet} in mmsegementation~\cite{mmseg}. See supplementary materials for more details.

\vspace{0.1cm}
\noindent\textbf{Results.} Table~\ref{table:seg} lists the parameters, FLOPs, FPS, single-scale (SS) and multi-scale (MS) mIoU for different backbones based on two typical frameworks. We first choose Semantic FPN~\cite{semanticfpn} as the basic framework following~\cite{pvt,cyclemlp}. It can be seen that Hire-MLP outperforms CycleMLP~\cite{cyclemlp} and PVT~\cite{pvt} by a large margin (44.3 vs. 42.4) with similar FLOPs and higher FPS, indicating the superiority of hierarchical rearrangement operation to model at various input scales. In addition, we follow~\cite{swin,asmlp} to validate our Hire-MLP based on another commonly used framework UperNet~\cite{upernet}. The proposed Hire-MLP achieves better MS mIoU compared to the state-of-the-art Swin Transformer~\cite{swin}, and is +1.6 mIoU higher than Swin-T on SS mIoU. It seems that Swin Transformer can obtain a larger improvement during multi-scale testing. We speculate one main reason is that the self-attention mechanism in Swin Transformer can capture scale information easier than our hire module. The related ablation studies can be found in supplementary materials.

\section{Conclusion}
This paper proposes a novel variant of MLP-based architecture via hierarchically rearranging tokens to aggregate both local and global spatial information. Input features are first split into multiple regions along the height/width directions. Different tokens in each region can communicate adequately via inner-region rearrangement operation, which mixes channels from different tokens to extract local information. Then tokens from different regions are rearranged by token shifting. This cross-region rearrangement operation not only exchanges the information between regions, but also preserves the relative position. Based on hierarchical rearrangement operations above, an effective and efficient Hire-MLP is constructed and has achieved significant performance improvements in various vision tasks.

{\small
\bibliographystyle{ieee_fullname}
\bibliography{egbib}
}

\end{document}